\documentclass[10pt,twocolumn]{article}

\usepackage[T1]{fontenc}
\usepackage[utf8]{inputenc}
\usepackage{lmodern}
\usepackage{microtype}
\usepackage[margin=0.72in]{geometry}
\usepackage{amsmath,amssymb}
\usepackage{bm}
\usepackage{booktabs}
\usepackage{tabularx}
\usepackage{array}
\usepackage{enumitem}
\usepackage{balance}
\usepackage[hidelinks]{hyperref}
\hypersetup{
  pdftitle={Evidence-Aware Reduction for Forkable Compute},
  pdfauthor={Yossi Eliaz},
  pdfsubject={Evidence-aware reduction, provenance, uncertainty, and forkable compute},
  pdfkeywords={distributed inference, evidence provenance, confidence distributions, forkable compute, uncertainty}
}

\setlist{leftmargin=1.3em,topsep=1pt,itemsep=0pt,parsep=0pt}
\usepackage{titlesec}
\titlespacing*{\section}{0pt}{1.1ex}{0.5ex}
\titlespacing*{\subsection}{0pt}{0.8ex}{0.4ex}
\titleformat{\section}{\normalfont\large\bfseries}{\thesection}{0.6em}{}
\titleformat{\subsection}{\normalfont\normalsize\bfseries}{\thesubsection}{0.5em}{}

\title{\vspace{-1.5em}\textbf{Evidence-Aware Reduction for Forkable Compute}}

\author{%
  Yossi Eliaz\\
  \normalsize Incredibuild \quad\textbar\quad islo.dev \quad\textbar\quad HIT CS Department, Israel\\
  \normalsize\texttt{yossi.eliaz@incredibuild.com} \quad \texttt{yossi@islo.dev} \quad \texttt{eliazy@hit.ac.il}
}
\date{}

\begin{document}
\maketitle

\begin{abstract}
Forkable runtimes execute multiple workers from a common prepared state. Runtime isolation does not
imply statistical independence: workers may share observations, evaluation data, model state,
prompts, tests, retrieved context, evaluators, or execution ancestry. A reducer that receives only
point outputs or scores cannot determine whether agreement reflects independent evidence or repeated
or correlated evidence. We introduce an \emph{evidence-aware reduction contract} in which each
worker returns an estimate, uncertainty, evidence identifiers, lineage, and execution metadata. For
independent workers estimating a common parameter, the reference implementation uses established
inverse-information pooling in Gaussian/Wald form. Its natural-parameter state merges associatively
through reduction trees; provenance follows separate semantics, and exact repeated nonempty evidence
identifiers are rejected by default. The same algebra retains a residual heterogeneity statistic
$\Delta$, equal to Cochran's $Q$ in the scalar inverse-variance case. Deterministic tests verify the
algebra, provenance transport, unequal-information behavior, and a forged-precision failure mode; a
named-snapshot trace exercises the worker-to-reducer path, while provider logs document three distinct
execution paths. The estimator is established. The contribution is the reduction contract and its
explicit separation of execution multiplicity from independent statistical evidence. Extending this
contract to dependence-aware reduction over fork DAGs remains open. Artifact:
\href{https://github.com/zozo123/boltzmann-mapreduce}{github.com/zozo123/boltzmann-mapreduce}.
\end{abstract}

\section{Introduction}
\label{sec:intro}

Forkable runtimes can replicate prepared state across many concurrent workers. These workers are
distinct executions, but their statistical evidence need not be distinct. Shared observations,
evaluation data, model state, prompts, tests, retrieved context, evaluators, or execution ancestry
can induce correlated error even when runtime isolation is strong. Large-scale measurements across
more than 350 language models provide a concrete example: model errors remain substantially
correlated across architectures and providers~\cite{kim2025correlated}.

The statistical consequence is immediate. If $K$ scalar estimates
have equal marginal variance $\sigma^2$ and exchangeable pairwise correlation $\rho$, then
\begin{equation}
\operatorname{Var}(\bar\theta)=
\sigma^2\!\left[\rho+\frac{1-\rho}{K}\right].
\label{eq:correlation-floor}
\end{equation}
An independence calculation instead reports $\sigma^2/K$. For fixed $\rho>0$, additional branches
cannot drive variance below $\rho\sigma^2$. Execution isolation controls runtime state; statistical
independence depends on what evidence and error sources the branches share.

Distributed systems make execution identity explicit, while statistical reduction often receives
only a score, vote, or candidate answer. Such outputs omit four quantities needed to justify a
precision claim: uncertainty, information strength, evidence identity, and inherited dependence.
MapReduce demonstrated the value of a small worker/reducer interface for placement, partitioning,
retries, and tree aggregation~\cite{dean2004mapreduce}. For forkable inference, the corresponding
interface question is what a worker must return so that reduction preserves the assumptions behind
its uncertainty estimate.

We define an evidence-aware worker record and a mergeable reduction state. The current artifact
implements the independent, common-target case using standard Gaussian/Wald pooling. The design
uses the following requirement:

\medskip
\noindent\textbf{Reduction invariant.}
\emph{A reducer may report additional precision only to the extent justified by its statistical
model of the evidence represented by the worker summaries. Worker count by itself carries no
statistical information.}
\medskip

This requirement affects ordinary runtime behavior. Retries and speculative duplicates should be
idempotent with respect to evidence identity. Numeric state and provenance require different merge
semantics. Shared ancestry should remain available after aggregation. When dependence is unresolved,
the reducer should retain that fact rather than silently applying an independence model.

\noindent\textbf{Contributions.}
\begin{itemize}
  \item \textbf{Reduction contract.} A worker result that couples an estimate and uncertainty with
  evidence identifiers, lineage, and execution metadata, including explicit semantics for exact
  evidence reuse, retries, and hierarchical aggregation.
  \item \textbf{Reference realization.} An associative Gaussian reducer that preserves the numeric
  state required for pooled uncertainty and heterogeneity while returning provenance with the final
  result, together with an auditable implementation and execution traces.
  \item \textbf{Statistical boundary.} A separation between established independent-evidence pooling
  and the unresolved problem of deriving dependence models from shared evidence, lineage, and
  adaptive selection.
\end{itemize}

\section{Evidence-Aware Worker Results}
\label{sec:contract}

\begin{figure*}[t]
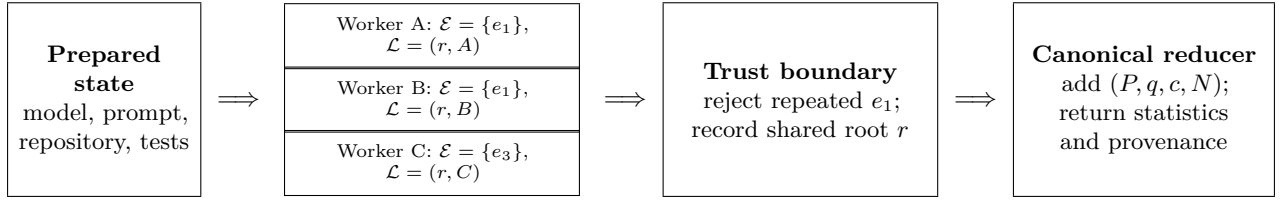

\centering
\setlength{\tabcolsep}{3pt}
\renewcommand{\arraystretch}{1.05}
\begin{tabular}{c c c c c c c}
\fbox{\parbox[c][0.92in][c]{0.13\textwidth}{\centering\small
\textbf{Prepared state}\\
model, prompt,\\repository, tests}}
& $\Longrightarrow$ &
\begin{tabular}{c}
\fbox{\parbox[c][0.25in][c]{0.205\textwidth}{\centering\scriptsize
Worker A: $\mathcal E=\{e_1\}$, $\mathcal L=(r,A)$}}\\[-1pt]
\fbox{\parbox[c][0.25in][c]{0.205\textwidth}{\centering\scriptsize
Worker B: $\mathcal E=\{e_1\}$, $\mathcal L=(r,B)$}}\\[-1pt]
\fbox{\parbox[c][0.25in][c]{0.205\textwidth}{\centering\scriptsize
Worker C: $\mathcal E=\{e_3\}$, $\mathcal L=(r,C)$}}
\end{tabular}
& $\Longrightarrow$ &
\fbox{\parbox[c][0.92in][c]{0.19\textwidth}{\centering\small
\textbf{Trust boundary}\\
reject repeated $e_1$;\\
record shared root $r$}}
& $\Longrightarrow$ &
\fbox{\parbox[c][0.92in][c]{0.18\textwidth}{\centering\small
\textbf{Canonical reducer}\\
add $(P,q,c,N)$;\\
return statistics\\and provenance}}
\end{tabular}
\caption{Execution multiplicity is not evidence multiplicity. Workers A and B declare the same
supporting evidence $e_1$, so the exact-overlap guard rejects their joint reduction. When evidence
identifiers differ, shared ancestry such as root $r$ remains visible for downstream dependence
modeling rather than being silently converted into independent precision.}
\label{fig:contract}
\end{figure*}

\paragraph{Worker record.}
Worker $k$ emits
\begin{equation}
r_k=(\hat\theta_k,J_k,n_k,\mathcal E_k,\mathcal L_k,m_k),
\label{eq:contract}
\end{equation}
where $\hat\theta_k\in\mathbb R^p$ is a local estimate, $J_k$ is estimated per-observation
information, $n_k$ is a literal sample count, $\mathcal E_k$ contains evidence identifiers,
$\mathcal L_k$ records fork lineage, and $m_k$ stores execution metadata. The reference dataclass
has a JSON-compatible serialized form. At ingestion it rejects non-finite estimates, non-positive
integer counts, malformed provenance, inconsistent dimensions, and non-symmetric or
non-positive-definite $J_k$.

The statistical invariant is total precision
\begin{equation}
P_k=n_kJ_k,\qquad q_k=P_k\hat\theta_k,\qquad
c_k=\hat\theta_k^TP_k\hat\theta_k.
\label{eq:canonical}
\end{equation}
The decomposition $P_k=n_kJ_k$ matches the current examples, where $n_k$ is a literal observation
count. More general workers may naturally report total precision or covariance directly, with
sample size or effective evidence mass represented separately. Likewise, positive-semidefinite
local information could be useful when different workers identify complementary directions; the
current implementation intentionally requires positive definiteness at the worker boundary.

\paragraph{Trust boundary.}
Evidence identifiers and lineage are declarations, not proofs. The reducer rejects exact overlap
among nonempty $\mathcal E_k$ sets unless the caller explicitly overrides the check. This catches
literal reuse such as processing the same evaluation token twice. It does not detect relabeled data,
partial overlap, common latent noise, shared evaluators, or two branches that make the same mistake
for the same hidden reason. Dimensions also do not establish a common estimand. A production system
should bind evidence, model, prompt, tool, evaluator, and target identities to authenticated or
content-addressed manifests before those records are used to justify precision.

Generic provenance standards such as W3C PROV provide interoperable representations of entities,
activities, agents, and derivation relationships~\cite{lebo2013prov}. Agent-provenance work is
increasingly making evidence, tool use, memory, and execution traces auditable
~\cite{wang2026traces}. Our narrower requirement is statistical: provenance must remain connected
to the uncertainty that a reducer reports.

\paragraph{Numeric merge and provenance merge.}
Independent summaries add
\begin{equation}
(P,q,c,N)=\bigoplus_k(P_k,q_k,c_k,n_k).
\label{eq:merge}
\end{equation}
The numeric state is associative and commutative in exact arithmetic, enabling flat, streaming,
or tree reduction. The scalar $c$ is essential: replacing an intermediate group by only its pooled
center and precision would discard disagreement accumulated inside that group. Provenance uses
separate semantics. Evidence IDs are canonicalized as a set; lineage uses order-preserving
deduplication and is therefore order-sensitive as serialized metadata. Finalization returns both
fields with the pooled statistics. Numeric state costs $O(p^2)$; an exact provenance ledger can
grow with the number of evidence and lineage entries.

\paragraph{Retries and speculative execution.}
An evaluation identity belongs to the evidence, not to the process that happened to execute it. A
retry or speculative duplicate should therefore reuse the same evidence token. Accepting both would
convert fault tolerance into double counting. The current reducer raises on repeated tokens; a
production keyed admission layer would normally choose one successful execution before reduction.
A branch that acquires genuinely new evidence receives a new token while retaining ancestry and
selection context.

\paragraph{When sufficient statistics are available.}
Model-specific sufficient statistics remain preferable when they can be transmitted: a mean from
$(\sum_i x_i,n)$ or OLS from $(X^TX,X^Ty)$. The present contract targets settings in which workers
naturally expose estimates and calibrated uncertainty while raw observations or model-specific
sufficient statistics are unavailable, expensive, private, or heterogeneous across worker types.

\section{Associative Gaussian Reference Reducer}
\label{sec:theory}

\paragraph{Statistical scope.}
Confidence distributions, inverse-information pooling, and distributed estimating-function methods
already establish how independent evidence can be combined~\cite{singh2005combining,xie2011meta,
xie2013confidence,zhou2017raocd,tang2020distributed}. We use the Gaussian/Wald common-effect
special case as a reference reducer rather than claiming a new estimator. Assume disjoint evidence,
independent estimation noise, a common parameter $\theta_0$, fixed $p$ and $K$,
$\min_k n_k\to\infty$, and calibrated positive-definite precision estimates. Let $V_k$ denote the
per-observation asymptotic covariance and suppose
\begin{equation}
n_k^{1/2}(\hat\theta_k-\theta_0)
\overset{d}{\longrightarrow}N_p(0,V_k),\qquad
J_k\overset{p}{\longrightarrow}V_k^{-1}.
\label{eq:an}
\end{equation}
For estimating equations, $V_k$ may be a sandwich covariance and $V_k^{-1}$ the corresponding
Godambe precision~\cite{godambe1960estimating}. The leading Wald kernel is
\begin{equation}
g_k(\theta)=\exp\!\left\{-\tfrac12
(\theta-\hat\theta_k)^TP_k(\theta-\hat\theta_k)\right\}.
\label{eq:factor}
\end{equation}
It is exact for a Gaussian local estimate with fixed known covariance and otherwise a first-order
approximation under the stated regularity conditions.

\paragraph{Pooling and heterogeneity.}
Let $P=\sum_kP_k$, $q=\sum_kq_k$, and $c=\sum_kc_k$. Completing the square gives
\begin{equation}
\hat\theta=P^{-1}q,\qquad \Sigma_g=P^{-1},
\label{eq:pool}
\end{equation}
where $\Sigma_g$ is the normalized product kernel's covariance and, under Eq.~\eqref{eq:an}, a
first-order sampling-covariance estimate for $\hat\theta$. The residual quadratic is
\begin{equation}
\begin{aligned}
\Delta
&=\sum_k(\hat\theta_k-\hat\theta)^TP_k(\hat\theta_k-\hat\theta)\\
&=c-q^TP^{-1}q\ge0
\quad\text{in exact arithmetic}.
\end{aligned}
\label{eq:heterogeneity}
\end{equation}
For $p=1$, $\Delta$ is Cochran's $Q$; the vector expression has direct multivariate meta-analytic
analogues~\cite{cochran1954combination,jackson2012heterogeneity}. Under independent
$N_p(\theta_0,P_k^{-1})$ estimates with fixed known $P_k$, $\Delta\sim\chi^2_{p(K-1)}$; plug-in
precision makes that calibration asymptotic. Importantly, small $\Delta$ does not establish
independence: correlated workers can agree because they share the same error.

For the unit-height kernels of Eq.~\eqref{eq:factor},
\begin{equation}
Z_g=(2\pi)^{p/2}|P|^{-1/2}\exp(-\Delta/2).
\label{eq:normalizer}
\end{equation}
Thus $\Delta/2$ is the heterogeneity contribution to $-\log Z_g$. The full integral also depends
on $|P|$, normalization conventions, and the integration measure; it should not be interpreted as
standalone model evidence or a calibrated conflict score.

\paragraph{Numerical implementation.}
The artifact uses Cholesky factorization for solves, covariance, and $\log|P|$. It clips a negative
computed $\Delta$ only within a scale-aware floating-point tolerance and raises on larger violations
of the canonical invariant. The implementation also exposes the equivalent Gibbs notation
$g_k=\exp\{-n_kE_k\}$ with
$E_k(\theta)=\tfrac12(\theta-\hat\theta_k)^TJ_k(\theta-\hat\theta_k)$. This is an interpretation,
not an additional statistical result: $P_k$ is invariant and $n_k$ may be absorbed into $J_k$.
When local sample sizes stay small or the number of shards grows too rapidly, local approximation
error and bias can dominate~\cite{zhang2013oneshot,zhou2017raocd}.

\paragraph{Reported precision is an attack surface.}
Holding reported precisions fixed, the local influence of worker $o$ is
\begin{equation}
\frac{\partial\hat\theta}{\partial\hat\theta_o}=P^{-1}P_o.
\label{eq:influence}
\end{equation}
A worker that can fabricate large precision can therefore dominate the result. The artifact includes
a transparent stress-test heuristic that caps determinant-based precision volume and applies a
coordinate-wise median/MAD redescending location weight. The determinant cap does not bound every
directional eigenvalue, and the location score is coordinate-dependent. It illustrates one failure
mode rather than providing a Byzantine-robustness guarantee.

\section{Evaluation: What the Artifact Establishes}
\label{sec:eval}

The evaluation intentionally separates algebraic verification, seeded behavioral checks, and
execution-path traces. All statistical data are synthetic. The evidence supports implementation
and interface claims; it is not a repeated-sampling study of statistical efficiency or a controlled
cross-provider systems benchmark.

\begin{table}[t]
\centering\footnotesize
\caption{Existing artifact checks and the claims they support. Logistic values are mean $\pm$ SD
over eight fixed seeds.}
\label{tab:checks}
\begin{tabularx}{\columnwidth}{@{}>{\raggedright\arraybackslash}p{0.22\columnwidth}>{\raggedright\arraybackslash}p{0.35\columnwidth}X@{}}
\toprule
Check & Observation & Supported conclusion \\
\midrule
Canonical algebra & Flat/tree and closed-form results agree & Implementation matches the Gaussian formulas \\
Evidence overlap & Repeated nonempty ID raises & Exact declared reuse is blocked \\
Unequal-size logistic & pool gap $0.0083\!\pm\!0.0042$; equal average $0.177\!\pm\!0.090$ & Information pooling beats unweighted averaging here \\
Forged precision & attacked $17.0004$; clipped $4.9566$ & One precision-fabrication failure mode and heuristic response \\
Snapshot integration & four workers complete in $6.70$\,s & Named-snapshot worker/reducer path executes end to end \\
\bottomrule
\end{tabularx}
\end{table}

\paragraph{Algebra and local estimators.}
Twelve homoscedastic mean shards of size $107$--$355$ produce pooled mean $4.9542$ with nominal
plug-in $95\%$ Wald interval $[4.8980,5.0104]$; the full-sample mean is $4.9546$. The reducer agrees
with closed-form inverse-variance pooling to floating-point precision, and flat and tree reductions
agree. This is an algebra check; sufficient-statistic reduction would recover the full sample mean
exactly. In the linear-regression check, the pooled coefficient has norm gap $0.033$ to centralized
OLS, with each worker estimating its own residual variance.

\paragraph{Nonlinear and forged-precision checks.}
Five IID logistic shards have sizes $\{60,120,400,2000,5000\}$. Across eight fixed seeds,
information pooling is $0.0083\pm0.0042$ from the centralized MLE, while an equal coefficient
average is $0.177\pm0.090$ away. This confirms the expected value of information weighting under
strongly unequal shard sizes; it does not substitute for comparisons with Rao-CD, surrogate
likelihood, sample-size weighting, or distributed Newton methods. In the forged-precision check,
one scalar worker uses a distant estimate from a 2,000-point shard and inflates its reported
precision by a further $50\times$. Unprotected pooling moves to $17.0004$; the stress heuristic
returns $4.9566$.

\paragraph{Execution traces.}
One integration trace begins from a named $141$\,MB islo snapshot and restores four sandboxes. The
seed-pinned workers return pooled mean $4.9422$ versus full-sample $4.9450$; four concurrent
restore--run--capture round trips complete in $6.70$\,s. This trace validates the complete
snapshot-to-worker-to-reducer path at the client-observed boundary.

\begin{table*}[t]
\centering\scriptsize
\caption{Platform-specific traces already present in the artifact. Operation boundaries differ;
these rows demonstrate exercised API paths rather than rank provider latency. Daytona and
Tensorlake percentiles cover sandbox creation, while islo covers restore--run--capture.}
\label{tab:platform}
\begin{tabularx}{\textwidth}{@{}l l >{\raggedright\arraybackslash}X r r r r r r@{}}
\toprule
Platform & State source & Measured operation & Client conc. & Batch & Success & $p_{50}$ & $p_{95}$ & Total wall \\
\midrule
Daytona & default environment & create & 8 & 1024 & 1024/1024 & 0.20\,s & 1.12\,s & 190.3\,s \\
Tensorlake & default environment & create & 8 & 256 & 256/256 & 3.44\,s & 9.00\,s & 387.3\,s \\
islo & named 141\,MB snapshot & restore + run + capture & 12 & 256 & 255/256 & 6.87\,s & 9.04\,s & 240.3\,s \\
\bottomrule
\end{tabularx}
\end{table*}

The batched provider logs in Table~\ref{tab:platform} record different API operations and different
client concurrency. Higher requested concurrency also encountered validation or capacity limits in
some exploratory runs. These observations are useful operational evidence, but they do not isolate
snapshot restore cost or support cross-platform performance ranking.

\section{Dependence-Aware Reduction}
\label{sec:agenda}

The reference reducer covers the case in which the independence model is specified. A general
fork-aware reducer requires additional dependence information.

\paragraph{Structured fork DAGs.}
A lineage union records common ancestors but neither reconstructs a fork DAG nor maps ancestry to
covariance. A stronger record would identify parents, immutable evidence manifests, model/prompt/tool
versions, evaluators and tests, retrieved artifacts, and branch-selection events. These relations can
be represented using established provenance concepts~\cite{lebo2013prov}, but the reducer additionally
needs semantics connecting them to statistical dependence.

\paragraph{Dependence and conservative abstention.}
Known overlap might induce shared latent factors, covariance blocks, clusters, or conservative
evidence groups. With an estimated joint covariance $\Omega$ for stacked worker estimates, a natural
next step is generalized least-squares-style fusion rather than the block-diagonal independence model
implicit in Eq.~\eqref{eq:pool}. The harder problem is obtaining trustworthy $\Omega$ from workflow
state. When the runtime cannot justify a dependence model, the reduction invariant implies a conservative behavior: retain the unresolved
dependency and do not report the narrower interval implied by independence.

\paragraph{Different targets and adaptive selection.}
The common-$\theta_0$ assumption is a common-effect model. Non-IID observations can still be pooled
when workers identify the same parameter and report valid uncertainty; genuinely different estimands
require site effects, meta-regression, random effects, or another explicit model. Selection creates a
separate dependency: if tests or verifiers are used both to choose surviving branches and to estimate
their confidence, naive post-selection uncertainty can be optimistic. Selection events therefore
belong in provenance, not merely in scheduler logs.

\paragraph{Confidence sources.}
For AI agents, verbal confidence is not automatically statistical precision. Useful precision must
be empirically tied to outcomes through held-out tests, repeated trials, calibrated evaluators,
independent verification, or another validated uncertainty model. Correlated-error evidence for
modern LLMs~\cite{kim2025correlated} makes this distinction especially important for ensembles of
nominally diverse agents.

\section{Related Work}
\label{sec:related}

\paragraph{Distributed inference.}
The Gaussian reference reducer is established common-effect meta-analysis and confidence-distribution
pooling~\cite{cochran1954combination,singh2005combining,xie2011meta,xie2013confidence}.
Rao-CD and distributed GLM inference provide closely related estimating-equation constructions
~\cite{zhou2017raocd,tang2020distributed}; one-shot averaging and surrogate likelihood study
neighboring communication--statistics tradeoffs~\cite{zhang2013oneshot,jordan2019csl}. Our
contribution sits before these estimators in the runtime stack: it makes the evidence and dependence
assumptions required by reduction part of the worker/reducer interface.

\paragraph{Provenance and agent traces.}
W3C PROV standardizes interoperable provenance relations among entities, activities, and agents
~\cite{lebo2013prov}. Recent agent-provenance work surveys how retrieved evidence, tools, memory,
environment observations, intermediate claims, and actions can be connected for audit and trust
~\cite{wang2026traces}. Our focus is complementary: a provenance edge matters to reduction when it
changes how much independent information the system may claim.

\paragraph{Inference-time diversity and correlated error.}
Self-consistency and multiagent debate exploit multiple generations to improve selection or reasoning
~\cite{wang2023selfconsistency,du2024multiagent}. Their benefits do not imply independent errors.
Large-scale evidence shows substantial error correlation across LLMs, including models with different
architectures and providers~\cite{kim2025correlated}. Recent multi-agent fact-verification work
controls decision risk by combining disagreement-sensitive scores with statistical calibration
~\cite{kostka2026uncertainty}. That approach operates at the decision layer; our focus is the runtime
record needed to determine which worker outputs may be treated as separate statistical evidence.

\paragraph{Forkable systems and robust aggregation.}
MapReduce established programmable reduction~\cite{dean2004mapreduce}; Firecracker, Catalyzer, REAP,
and Faasm reduce the cost of isolated or reusable execution state
~\cite{agache2020firecracker,du2020catalyzer,ustiugov2021reap,shillaker2020faasm}. These systems make
fan-out practical. Robust distributed-learning methods such as Krum and FLAME instead begin from
explicit adversary models~\cite{blanchard2017krum,nguyen2022flame}. The current artifact has neither
a Byzantine guarantee nor a correlation estimator; it provides the evidence-carrying boundary on
which such reducers can be selected or built.

\section{Conclusion}
\label{sec:conclusion}

Forkable runtimes can increase the number of executions without increasing the amount of independent
evidence. A reduction interface that receives only worker outputs cannot distinguish independent
measurements from repeated or correlated evidence.

Evidence-aware reduction makes the missing state explicit. Workers return estimates together with
uncertainty, evidence identifiers, lineage, and execution metadata. For independent Gaussian
summaries, the resulting numeric state supports associative hierarchical reduction while retaining
heterogeneity and provenance. The reference implementation validates this contract across local and
snapshot-backed execution paths and rejects exact declared evidence reuse by default.

The remaining inference problem is to derive a defensible dependence model from shared evidence,
lineage, and selection history. Until that model is available, execution multiplicity should not be
interpreted as independent statistical information.

\balance
\bibliographystyle{abbrv}
\bibliography{refs}

@inproceedings{dean2004mapreduce,
  author    = {Dean, Jeffrey and Ghemawat, Sanjay},
  title     = {{MapReduce}: Simplified Data Processing on Large Clusters},
  booktitle = {Proceedings of the 6th USENIX Symposium on Operating Systems Design and Implementation (OSDI)},
  pages     = {137--150},
  year      = {2004}
}

@article{singh2005combining,
  author  = {Singh, Kesar and Xie, Min-ge and Strawderman, William E.},
  title   = {Combining Information from Independent Sources through Confidence Distributions},
  journal = {The Annals of Statistics},
  volume  = {33},
  number  = {1},
  pages   = {159--183},
  year    = {2005}
}

@article{xie2011meta,
  author  = {Xie, Minge and Singh, Kesar and Strawderman, William E.},
  title   = {Confidence Distributions and a Unifying Framework for Meta-Analysis},
  journal = {Journal of the American Statistical Association},
  volume  = {106},
  number  = {493},
  pages   = {320--333},
  year    = {2011},
  doi     = {10.1198/jasa.2011.tm09803}
}

@article{xie2013confidence,
  author  = {Xie, Min-ge and Singh, Kesar},
  title   = {Confidence Distribution, the Frequentist Distribution Estimator of a Parameter: A Review},
  journal = {International Statistical Review},
  volume  = {81},
  number  = {1},
  pages   = {3--39},
  year    = {2013},
  doi     = {10.1111/insr.12000}
}

@article{tang2020distributed,
  author  = {Tang, Lu and Zhou, Ling and Song, Peter X.-K.},
  title   = {Distributed Simultaneous Inference in Generalized Linear Models via Confidence Distribution},
  journal = {Journal of Multivariate Analysis},
  volume  = {176},
  pages   = {104567},
  year    = {2020},
  doi     = {10.1016/j.jmva.2019.104567}
}

@article{zhou2017raocd,
  author  = {Zhou, Ling and Song, Peter X.-K.},
  title   = {Scalable and Efficient Statistical Inference with Estimating Functions in the {MapReduce} Paradigm for Big Data},
  journal = {arXiv preprint arXiv:1709.04389},
  year    = {2017},
  url     = {https://arxiv.org/abs/1709.04389}
}

@article{godambe1960estimating,
  author  = {Godambe, V. P.},
  title   = {An Optimum Property of Regular Maximum Likelihood Estimation},
  journal = {The Annals of Mathematical Statistics},
  volume  = {31},
  number  = {4},
  pages   = {1208--1211},
  year    = {1960}
}

@article{cochran1954combination,
  author  = {Cochran, William G.},
  title   = {The Combination of Estimates from Different Experiments},
  journal = {Biometrics},
  volume  = {10},
  number  = {1},
  pages   = {101--129},
  year    = {1954},
  doi     = {10.2307/3001666}
}

@article{jackson2012heterogeneity,
  author  = {Jackson, Dan and White, Ian R. and Riley, Richard D.},
  title   = {Quantifying the Impact of Between-Study Heterogeneity in Multivariate Meta-Analyses},
  journal = {Statistics in Medicine},
  volume  = {31},
  number  = {29},
  pages   = {3805--3820},
  year    = {2012},
  doi     = {10.1002/sim.5453}
}

@article{zhang2013oneshot,
  author  = {Zhang, Yuchen and Duchi, John C. and Wainwright, Martin J.},
  title   = {Communication-Efficient Algorithms for Statistical Optimization},
  journal = {Journal of Machine Learning Research},
  volume  = {14},
  number  = {104},
  pages   = {3321--3363},
  year    = {2013},
  url     = {https://jmlr.org/papers/v14/zhang13b.html}
}

@article{jordan2019csl,
  author  = {Jordan, Michael I. and Lee, Jason D. and Yang, Yun},
  title   = {Communication-Efficient Distributed Statistical Inference},
  journal = {Journal of the American Statistical Association},
  volume  = {114},
  number  = {526},
  pages   = {668--681},
  year    = {2019},
  doi     = {10.1080/01621459.2018.1429274}
}

@inproceedings{agache2020firecracker,
  author    = {Agache, Alexandru and Brooker, Marc and Florescu, Andreea and Iordache, Alexandra and Liguori, Anthony and Neugebauer, Rolf and Piwonka, Phil and Popa, Diana-Maria},
  title     = {Firecracker: Lightweight Virtualization for Serverless Applications},
  booktitle = {Proceedings of the 17th USENIX Symposium on Networked Systems Design and Implementation (NSDI)},
  pages     = {419--434},
  year      = {2020}
}

@inproceedings{du2020catalyzer,
  author    = {Du, Dong and Yu, Tianyi and Xia, Yubin and Zang, Binyu and Yan, Guanglu and Qin, Chenggang and Wu, Qixuan and Chen, Haibo},
  title     = {Catalyzer: Sub-Millisecond Startup for Serverless Computing with Initialization-Less Booting},
  booktitle = {Proceedings of the 25th International Conference on Architectural Support for Programming Languages and Operating Systems (ASPLOS)},
  pages     = {467--481},
  year      = {2020}
}

@inproceedings{ustiugov2021reap,
  author    = {Ustiugov, Dmitrii and Petrov, Plamen and Kogias, Marios and Bugnion, Edouard and Grot, Boris},
  title     = {Benchmarking, Analysis, and Optimization of Serverless Function Snapshots},
  booktitle = {Proceedings of the 26th International Conference on Architectural Support for Programming Languages and Operating Systems (ASPLOS)},
  pages     = {559--572},
  year      = {2021}
}

@inproceedings{shillaker2020faasm,
  author    = {Shillaker, Simon and Pietzuch, Peter},
  title     = {Faasm: Lightweight Isolation for Efficient Stateful Serverless Computing},
  booktitle = {2020 USENIX Annual Technical Conference (USENIX ATC)},
  pages     = {419--433},
  year      = {2020}
}

@inproceedings{blanchard2017krum,
  author    = {Blanchard, Peva and El Mhamdi, El Mahdi and Guerraoui, Rachid and Stainer, Julien},
  title     = {Machine Learning with Adversaries: Byzantine Tolerant Gradient Descent},
  booktitle = {Advances in Neural Information Processing Systems},
  volume    = {30},
  pages     = {119--129},
  year      = {2017}
}

@inproceedings{nguyen2022flame,
  author    = {Nguyen, Thien Duc and Rieger, Phillip and Chen, Huili and Yalame, Hossein and M{\"o}llering, Helen and Fereidooni, Hossein and Marchal, Samuel and Miettinen, Markus and Mirhoseini, Azalia and Zeitouni, Shaza and Koushanfar, Farinaz and Sadeghi, Ahmad-Reza and Schneider, Thomas},
  title     = {{FLAME}: Taming Backdoors in Federated Learning},
  booktitle = {Proceedings of the 31st USENIX Security Symposium (USENIX Security)},
  pages     = {1415--1432},
  year      = {2022}
}

@inproceedings{wang2023selfconsistency,
  author    = {Wang, Xuezhi and Wei, Jason and Schuurmans, Dale and Le, Quoc V. and Chi, Ed H. and Narang, Sharan and Chowdhery, Aakanksha and Zhou, Denny},
  title     = {Self-Consistency Improves Chain of Thought Reasoning in Language Models},
  booktitle = {The Eleventh International Conference on Learning Representations},
  year      = {2023},
  url       = {https://openreview.net/forum?id=1PL1NIMMrw}
}

@inproceedings{du2024multiagent,
  author    = {Du, Yilun and Li, Shuang and Torralba, Antonio and Tenenbaum, Joshua B. and Mordatch, Igor},
  title     = {Improving Factuality and Reasoning in Language Models through Multiagent Debate},
  booktitle = {Proceedings of the 41st International Conference on Machine Learning},
  series    = {Proceedings of Machine Learning Research},
  volume    = {235},
  pages     = {11733--11763},
  publisher = {PMLR},
  year      = {2024},
  url       = {https://proceedings.mlr.press/v235/du24e.html}
}

@techreport{lebo2013prov,
  author      = {Lebo, Timothy and Sahoo, Satya and McGuinness, Deborah},
  title       = {{PROV-O}: The {PROV} Ontology},
  institution = {World Wide Web Consortium},
  type        = {W3C Recommendation},
  year        = {2013},
  month       = apr,
  url         = {https://www.w3.org/TR/prov-o/}
}

@inproceedings{kim2025correlated,
  author    = {Kim, Elliot Myunghoon and Garg, Avi and Peng, Kenny and Garg, Nikhil},
  title     = {Correlated Errors in Large Language Models},
  booktitle = {Proceedings of the 42nd International Conference on Machine Learning},
  series    = {Proceedings of Machine Learning Research},
  volume    = {267},
  pages     = {30038--30066},
  publisher = {PMLR},
  year      = {2025},
  url       = {https://proceedings.mlr.press/v267/kim25e.html}
}

@article{wang2026traces,
  author  = {Wang, Yiqi and Zhang, Jiaqi and Cai, Taotao and Liu, Zirui and Sun, Qingqiang and Sun, Zequn and Wu, Zhangkai and Zhang, Mingkai and Zhu, Yanming},
  title   = {From Agent Traces to Trust: Evidence Tracing and Execution Provenance in {LLM} Agents},
  journal = {arXiv preprint arXiv:2606.04990},
  year    = {2026},
  url     = {https://arxiv.org/abs/2606.04990}
}

@inproceedings{kostka2026uncertainty,
  author    = {Kostka, Adam and Chudziak, Jaroslaw A.},
  title     = {Controlling Uncertainty and Hallucination Risk in Multi-Agent Fact Verification},
  booktitle = {Proceedings of the 42nd Conference on Uncertainty in Artificial Intelligence},
  series    = {Proceedings of Machine Learning Research},
  volume    = {337},
  pages     = {3143--3161},
  publisher = {PMLR},
  year      = {2026},
  url       = {https://proceedings.mlr.press/v337/kostka26a.html}
}

\end{document}